\documentclass[10pt,twocolumn,letterpaper]{article}

\usepackage[pagenumbers]{wacv} 

\usepackage{graphicx}
\usepackage{amsmath}
\usepackage{amssymb}
\usepackage{booktabs}
\usepackage{enumitem}

\usepackage[pagebackref,breaklinks,colorlinks]{hyperref}

\usepackage[capitalize]{cleveref}
\crefname{section}{Sec.}{Secs.}
\Crefname{section}{Section}{Sections}
\Crefname{table}{Table}{Tables}
\crefname{table}{Tab.}{Tabs.}

\def\wacvPaperID{*****} 
\def\confName{WACV}
\def\confYear{2025}

\begin{document}

\title{A Mamba-based Siamese Network for Remote Sensing Change Detection}

\author{Jay N. Paranjape\\
Johns Hopkins University, Baltimore\\
jparanj1@jhu.edu\\
\and
Celso de Melo\\
DEVCOM Army Research Laboratory, Adelphi\\
celso.m.demelo.civ@army.mil\\
\and
Vishal M. Patel\\
Johns Hopkins University, Baltimore\\
vpatel36@jhu.edu\\
}
\maketitle

\begin{abstract}
   Change detection in remote sensing images is an essential tool for analyzing a region at different times. It finds varied applications in monitoring environmental changes, man-made changes as well as corresponding decision-making and prediction of future trends. Deep learning methods like Convolutional Neural Networks (CNNs) and Transformers have achieved remarkable success in detecting significant changes, given two images at different times. In this paper, we propose a {\bf{M}}amba-based {\bf{C}}hange {\bf{D}}etector {\bf{(M-CD)}} that segments out the regions of interest even better. Mamba-based architectures demonstrate linear-time training capabilities and an improved receptive field over transformers. Our experiments on four widely used change detection datasets demonstrate significant improvements over existing state-of-the-art (SOTA) methods. Our code and pre-trained models are available at \url{https://github.com/JayParanjape/M-CD}
\end{abstract}

\section{Introduction}
\label{sec:intro}
In remote sensing, Change Detection (CD) refers to the task of detecting significant changes to our planet's surface across time. It is generally addressed by processing satellite images of a region, taken at different instances in time \cite{cd1,cd2}. The task of CD is non-trivial since the change of interest can vary across different applications, ranging from man-made structures, natural vegetation, or effects of climate change. Subsequently, CD finds its applications in monitoring and assessing natural disasters and climate shifts \cite{climateshift_cd1,climateshift_cd2,disaster_cd1,disaster_cd2,disaster_cd3}, policy planning \cite{policy_cd1,policy_cd2}, surveying land and farmland cover \cite{land_cd1,land_cd2,land_cd3}, as well as military applications \cite{military_cd1,military_cd2}. Remote sensing images over different points in time can have multiple data disparities, including illumination changes \cite{illumination_challenge1,illumination_challenge2}, varying resolutions \cite{resolution_challenge1}, noise \cite{noise_challenge1, noise_challenge2}, and errors in registration \cite{registration_challenge1, registration_challenge2} among others. Thus, making CD systems that are robust to such discrepancies is a complex and challenging task.

With the advent of deep learning, various systems have been developed for CD that employ deep networks and show substantial gains on multiple datasets \cite{snunet,dtscn,bit,changeformer, ddpmcd, ifnet}. These methods include Convolutional Neural Networks \cite{snunet,fcsiam,dtscn}, Attention-based networks \cite{changeformer,bit,stanet} as well as diffusion-based networks \cite{ddpmcd}. Deep learning-based methods for CD generally take in a pair, consisting of a pre-change image and a post-change image as input, and output a segmentation mask locating the regions where the change of interest has occurred.  

\begin{figure}[t!]
  \center
  {\includegraphics[width=0.85\linewidth]{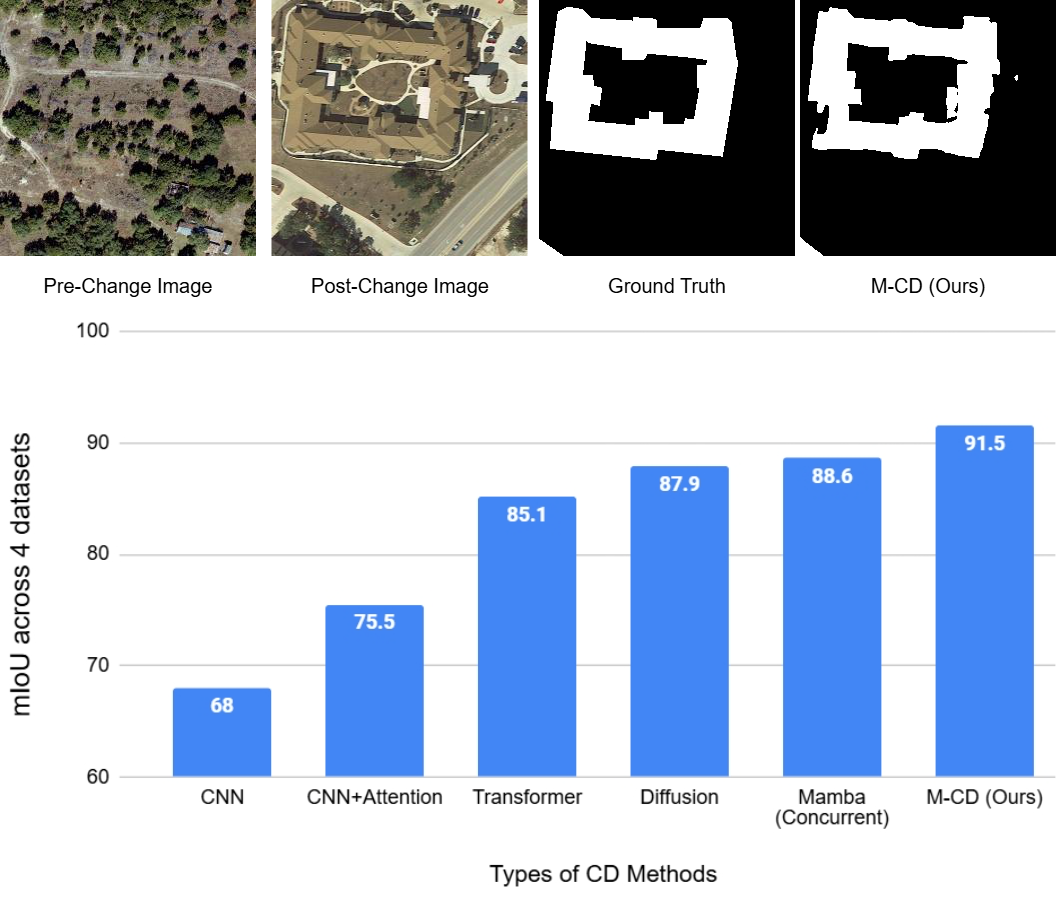}}
\caption{(Top) Example result of our M-CD. (Bottom) Average performance of M-CD with respect to existing types of CD models}  \label{fig:main_pg1_fig}
\end{figure}

Recently, selective state space models (a.k.a Mamba) \cite{mamba} were proposed that have a more global receptive field than transformers and are more scalable. While these were introduced for text inputs, VMamba \cite{vmamba} adapted this model for representing images and showed improvements in general vision tasks like classification \cite{vmamba} and segmentation \cite{sigma}. In this paper, we propose a Mamba-based Change Detector (M-CD) that outperforms all existing SOTA CD methods, as shown in \cref{fig:main_pg1_fig}. In particular, we develop a VMamba-based Siamese image encoder that is used to generate rich features from the two input images. We then develop a concatenation-based difference module that takes the features for the two images as input and combines them at multiple scales to create a joint feature vector. This is passed to a VMamba-based decoder that performs an aggregation over channels to produce the final output mask. In summary, the main contributions of our work are as follows:
\begin{enumerate}[noitemsep]
    \item We develop a novel Mamba-based Change Detector, called M-CD, that generates precise change masks, given two images.
    \item To this extent, we develop a difference module that combines the features from the pair of images. This is provided to the mask decoder, which is modified to be more aware of the channel information along with the temporal and spatial information.
    \item We evaluate our method on four widely used public datasets. We demonstrate consistent improvements over existing SOTA methods.
\end{enumerate}


\section{Related Work}
\noindent {\bf{Classical Change Detection.}}
Traditionally, CD has been approached using algebraic, transformation-based or classification-based methods. Algebraic methods generally perform predefined mathematical operations to find a difference measure between the pre-change and post-change images. This is followed by using a pre-defined threshold to detect changed areas in the image. Some examples of these methods include ImageDiff \cite{alge1}, ImageRegr \cite{alge2}, ImageRatio \cite{alge1}, and Change Vector Analysis (CVA) \cite{alge3}. While these methods are simple to implement, they are not sufficient for modeling the complexity of the problem and are heavily reliant on the threshold.

Transformation-based methods generally apply various transforms to the images and then compute a ``difference" image based on the transformed input images. Examples of these transforms include Principal Component Analysis \cite{pca1,pca2}, Karhunen-Loeve Transform (KT) \cite{alge1}, Gramm-Schmidt (GS) \cite{alge1}, Multivariate Alteration Detection (MAD) \cite{mad}, Re-Weighted Multivariate Alteration Detection (IRMAD) \cite{irmad}, and Chi-Square \cite{alge1}. The common principle behind these methods is that they aim to condense the important information present in the input pair of images using transforms. With just the relevant information remaining after the transform, they propose that a difference operation along with a threshold can give the required segmentation map. However, these methods are also limited by the selection of threshold as well as poor performance.

Finally, classification-based methods like Expectation-Maximization CD \cite{alge1} and spectral-temporal CD \cite{alge1} use machine learning techniques to provide per-pixel classification and perform additional post-processing to generate the final mask. However, as compared to modern-day deep networks, these techniques have an inferior quality.\\

\noindent {\bf{Change Detection Using Deep Learning.}}
Deep learning techniques produce rich embeddings for images, allowing them to surpass traditional methods for CD. Convolutional Neural Networks paved the way for utilization of DL for processing the pre and post-change images and then fusing them to produce the predicted mask. FC-EF \cite{fcsiam} and FC-Siam-Conc \cite{fcsiam} were one of the first methods to do so. In FC-EF, the pre-change and post-change image are concatenated along the channel dimension and a fully convolutional encoder-decoder structure is applied on the joint image. In FC-Siam-Conc, the same encoder operates on both the input images and the resulting embeddings are concatenated before passing them through the decoder. The latter approach was shown to be more effective in modeling the temporal difference between the two images. Later, with the introduction of deeper backbones like VGG \cite{vgg}, Resnet \cite{resnet} and DenseNet \cite{densenet}, multiple CD techniques have evolved which improve the performance of FC-Siam-Conac. Some examples include DT-SCN \cite{dtscn} and SNUNet \cite{snunet}. DT-SCN combines three subnetworks, one for change detection and two for semantic segmentation, which adopt a ResNet backbone. SNUNet combines dense connections and nested UNet \cite{unetpp} with Siamese networks for CD.

The introduction of Transformer networks for image analysis tasks in ViT \cite{vit} was instrumental in capturing the long-range dependencies in images, which was previously missing from CNN architectures. Since CD involves analyzing the relation between two different images, transformer-based approaches like BIT \cite{bit} and ChangeFormer \cite{changeformer} were found to be more effective than CNN methods \cite{bit,changeformer}. BIT uses a combination of ResNet, followed by a transformer encoder, that allows it to extract spatial and comparative features. Its successor, the Changeformer uses a purely transformer architecture and showed significant improvements over BIT. It comprises of a transformer encoder for the images as well as a transformer decoder, which is more effective than CNNs for feature representations.

One recent line of work uses self-supervised learning to pretrain the models with vast amounts of unannotated data. These may be obtained from datasets like ImageNet 1k as well as publicly available data on the internet. Some examples of this approach include SiamSiam \cite{siamsiam}, MoCo-v2 \cite{mocov2}, DenseCL \cite{densecl}, CMC \cite{cmc}, SeCo \cite{seco} and SaDL-CD \cite{sadlcd}, among others. These methods usually pretrain their encoders on self-supervision tasks, thereby vastly improving their representative nature. For instance, SaDL-CD performs background-swapping during training and applies data augmentations to produce three different views of a given image. Next, it optimizes a loss function that appropriately measures the difference between the views. SeCo, on the other hand, enforces consistencies between features of the seasonally varying images.

Another recent work proposes the use of a generative modeling framework called Denoising Diffusion Probabilistic Models (DDPM) \cite{ddpm} as a pretraining strategy for learning powerful encoders \cite{ddpmcd}. Once learnt, they can be used as feature extractors to train a light-weight model for CD. Hence, these can also be considered as a method that requires a large amount of pretraining. The proposed approach, called DDPM-CD, outperforms all the traditional and deep learning techniques mentioned above and is the current SOTA method for CD.

In this paper, we develop a Mamba-based mechanism that outperforms DDPM-CD as well as other methods on four widely used public datasets for CD, showing the effectiveness of selective state space modeling as a fundamental building block for CD over CNNs and transformers.\\

\noindent {\bf{Mamba in Vision Tasks.}}
Various recent works propose the use of State Space Modeling (S4) as an effective method of representing visual information \cite{s4nd,vis4mer,selective,trans4mer}. Trans4mer \cite{trans4mer} and Vis4mer \cite{vis4mer} were used to create representations of movie clips using S4, while S4ND \cite{s4nd} uses S4 to effectively model multidimensional signals in 1D, 2D and 3D. On the other hand, S5 \cite{selective} proposes a selective mechanism for capturing long-range dependencies in long videos, which became the motivation for Mamba \cite{mamba}. 
Mamba has been adapted for a multitude of computer vision tasks recently, which shows its effectiveness in modeling visual representations. Vision Mamba \cite{visionmamba} proposed using the selective scan mechanism in Mamba from two directions for an image, while VMamba \cite{vmamba} showed that scanning an image from four directions has a better representation. In our work, we follow the four-directional strategy of VMamba in the Siamese Image Encoder. Mamba has also been explored for various applications in Computer Vision including medical image segmentation \cite{umamba,vmunet,weakmambaunet,segmamba}, multi-modal segmentation \cite{sigma}, point cloud computations \cite{pointmamba}, and image restoration \cite{mambair}. However, most of these methods involve direct replacement of transformer blocks with Mamba blocks, without adapting for the particular task. Unlike this plug-and-play approach, we develop a Mamba-based difference module and decoder that is tailored for the task of CD.\\

\noindent {\bf{Concurrent Works in Mamba for CD.}}
A couple of recent CD methods have proposed using Mamba as the backbone architecture. 
While working on the proposed method, we recognize these recent methods as concurrent works and compare our method with them. ChangeMamba \cite{changemamba} uses three different designs in the decoder for modeling the spatio-temporal relationship between the pre-change and post-change images. RSMamba \cite{rsmamba}, on the other hand increases the number of scanning directions in VMama from four to eight in order to capture relations from all directions. In contrast, our method uses a single decoder design and the standard four directions for scanning. Instead, we design the decoder to use features at multiple scales for spatial learning and design a separate difference module to learn temporal relations. CDMamba \cite{cdmamba} comes close to our proposed method in that it uses multi-scale features in the decoder. However, our method also additionally uses a difference module to allow the model to learn temporal relations between the two images better. Experimental results on multiple datasets show that our method outperforms concurrent Mamba-based methods showing the significance of our architecture design for CD.

\section{Proposed Method}

\subsection{Preliminaries}
\noindent\textbf{State Space Modeling. } State Space Models (SSM) are a class of sequence-to-sequence modeling algorithms, which follow the linear time invariance (LTI) property \cite{zoh,ssm2,ssm3}. These systems work by maintaining a hidden state that changes with time. The output of the system at any point in time is dependent on the hidden state as well as the input. More specifically, given input \(x(t) \in \mathbb{R}\), the following equations define the output \(y(t) \in \mathbb{R}\) and the hidden state \(h(t) \in \mathbb{R}^{N}\), where \(N\) is the size of the hidden state:
\begin{equation}
    y(t) = Ch(t) + Dx(t), \;\;\; \dot{h}(t) = Ah(t) + Bx(t),
\end{equation}
where \(\dot{h}(t)\) denotes the first derivative with respect to time \(t\). The matrices \(A \in \mathbb{R}^{NXN}\), \(B \in \mathbb{R}^{NX1}\), \(C \in \mathbb{R}^{1XN}\), and \(D \in \mathbb{R}\) are characteristic to the system and are learnt during training. Further, to model discrete sequences like images, SSMs utilize a predefined discretization parameter \(\Delta\), which maps the parameters \(A\) and \(B\) to the discrete space. This technique is known as Zero Order Hold Discretization \cite{zoh}. More specifically, the discretization process is defined as follows:
\begin{equation}
    y_k = \overline{C}h_k + \overline{D}x_k, \;\; h_k = \overline{A}h_{k-1}+\overline{B}x_k,
\end{equation}
where \(\{x_1,x_2,...,x_k\}\) denotes the discrete input sequence, \(\{y_1,y_2,...,y_k\}\) denotes the output sequence, and the SSM matrices \(\overline{A}, \overline{B}, \overline{C}, \overline{D}\) are defined as follows:
\begin{gather}
    \nonumber \overline{A} = exp(\Delta A), \;\; \overline{B} = (\Delta A)^{-1}(exp(A)-I) (\Delta B) \\
    \overline{C} = C, \;\; \overline{D} = D.
\end{gather}

\begin{figure*}
  \center
  {\includegraphics[width=\linewidth]{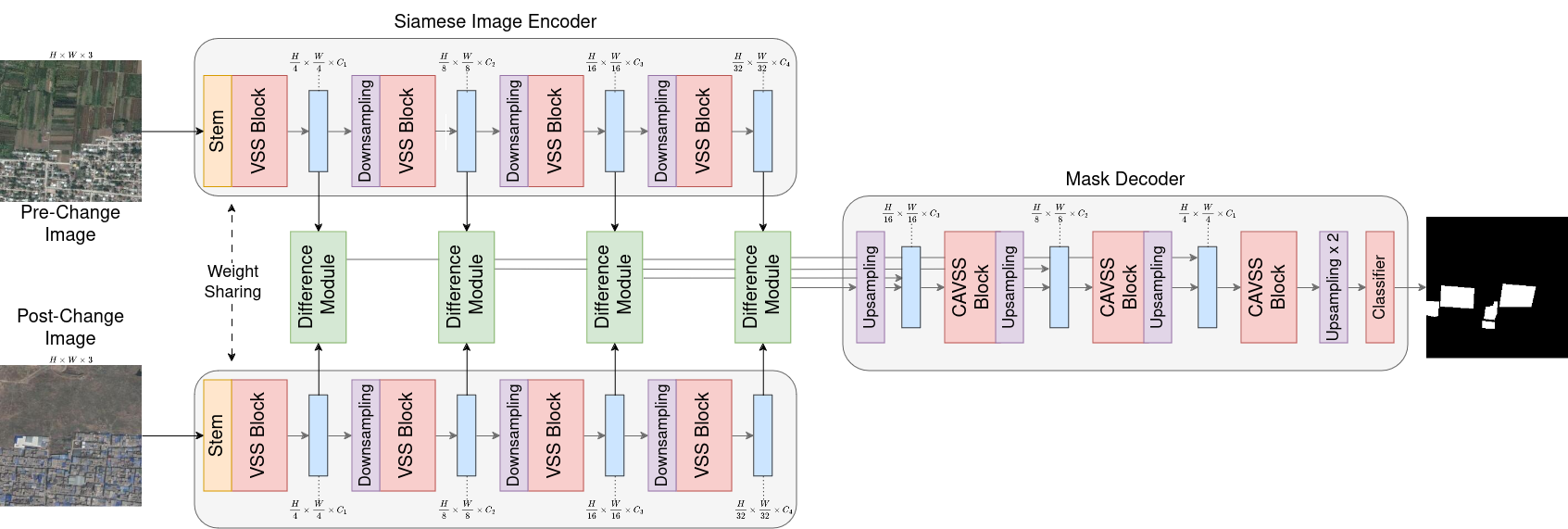}}
\caption{An overview of the M-CD Architecture. Given a pre-change and post-change image, they are passed through a Mamba-based encoder with shared weights (Siamese) and features at multiple scales are extracted. The Difference Module combines them before sending to the Mask Decoder, which uses a Mamba-based Decoder to generate the predicted change map.}  \label{arch}
\end{figure*}
\noindent\textbf{Selective Scan Modification. }The LTI property of SSMs causes them to learn a general set of parameters that are expected to work well for all kinds of inputs. Mamba (Selective State Space Model) \cite{mamba} addresses this concern by making the parameters input-dependent and shows that this allows the model to perform significantly better. More specifically, the parameters \(B, C\) and \(\Delta\) are dependent on the input \(x\) during training, making Mamba capable of effectively learning the complex relations in longer sequences. In addition, Mamba also approximates \(\overline{B} \simeq \Delta B\) using Taylor approximation.

\subsection{M-CD Architecture}
An overview of the proposed M-CD architecture is shown in Figure~\cref{arch}. M-CD consists of three main components - the Siamese Image Encoder (SIE), the Difference Module (DM) and the Mask Decoder (MD). Given two images, they are passed through the two branches of the encoder to generate image features. The two branches work on the same modality of images and so weights between them are shared. This also reduces the computational complexity. The SIE is responsible for extracting features on multiple scales, facilitated by the cascading of four Visual State Space (VSS) blocks and downsampling operations. The DM is responsible for analyzing features from both images together at different scales and generating combined multi-scale features. These are further transformed by the Mask Decoder using Channel-Averaged VSS blocks and upsampling operations. These transformed features are finally passed through a classifier that segments out the regions involving significant change. In the subsequent subsections, we describe each of the three components of M-CD in greater detail.

\subsection{Siamese Image Encoder (SIE)}
\begin{figure}[htp!]
  \center
  {\includegraphics[width=0.8\linewidth]{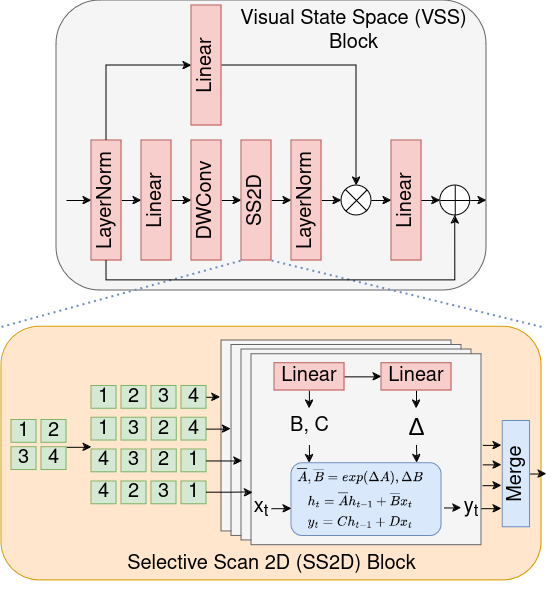}}
\caption{Architecture of Visual State Space Block and Selective Scan 2D block.}  \label{sie_arch}
\end{figure}

The SIE has two branches with common weights, each of which takes an RGB image. One of these is the pre-change image and the other is the post-change image. These images go through a stem module which is a series of 2D convolution layers to extract preliminary features, similar to Vision Transformer (ViT) \cite{vit}. This is followed by the VSS block for further processing. As seen in \cref{sie_arch}, the VSS block consists of a linear layer, followed by a Depth-wise Convolution layer, similar to the original Mamba \cite{mamba}. This is followed by the Selective Scan 2D (SS2D) Module, similar to VMamba \cite{vmamba} and MambaIR \cite{mambair}. As shown in \cref{sie_arch}, The SS2D module flattens the input and processes it from four different directions - top-left to bottom-right, bottom-right to top-left, top-right to bottom-left, and bottom-left to top-right. This is done to extract long-range dependencies from multiple directions, as suggested by VMamba \cite{vmamba}. For each of the directions, the selective scan state model is learnt by learning the parameters \(A, B, C, D, \Delta\), and the results are merged to produce the final output of the SS2D block. Lastly, the VSS block consists of a layernorm operation followed by a linear operator. The SIE consists of four such VSS blocks. After the first block, each of the VSS blocks are preceeded by a downsampling module. Thus, the SIE produces deep multi-scale features as the outputs of each of the VSS blocks for both the input images.

\subsection{Difference Module (DM)}
\begin{figure}[htp!]
  \center
  {\includegraphics[width=0.8\linewidth]{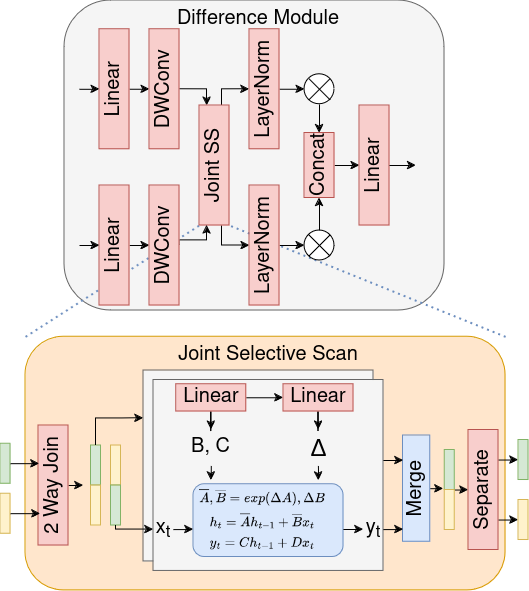}}
\caption{Architecture of the Difference Module and Joint Selective Scan.}  \label{dm_arch}
\end{figure}

As seen in \cref{dm_arch}, the difference module takes in features corresponding to the two images (pre-change and post-change), and outputs one feature vector to be used by the mask decoder. Each of the input features passes through a linear and a depth-wise convolution operation, followed by a Joint Selective Scan (JSS) module. The JSS module concatenates the outputs from the previous layer in two ways - \(Pre;Post\) and \(Post;Pre\), where \(Pre\) denotes the pre-change image, \(Post\) denotes the post-change image, and \(;\) represents concatenation. The selective scan operations are performed on these two concatenated features and the outputs are added. Unlike transformers which use self-attention to attend to each token, the DM uses a scanning operation. Hence, it is important to concatenate in both directions to maintain symmetry of the network and aid the training process. The resulting vector is then split into the pre-image and post-image parts and passed through a layernorm operation. The linear and convolution operations are used to refine the incoming features, while the JSS module is used for learning to identify the significant differences between the two images. In addition, a residual connection is added between the features before the convolution and after the normalization layer to aid training by reducing the effect of vanishing gradients. The resulting vectors are concatenated and passed through a final linear layer, before passing the result to the mask decoder. For capturing the difference information between images at different scales, we employ four separate DMs corresponding to four different scaled inputs from the image encoder.

\subsection{Mask Decoder (MD)}
\begin{figure}[htp!]
  \center
  {\includegraphics[width=0.7\linewidth]{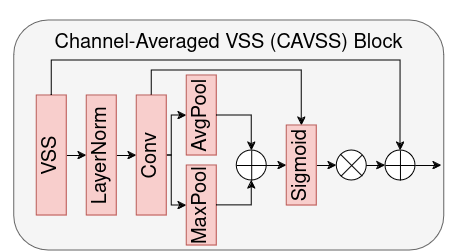}}
\caption{Architecture of Channel Averaged VSS Block}  \label{cvass_arch}
\end{figure}

As shown in \cref{arch}, the Mask Decoder is responsible for generating the output mask. The Mask decoder consists of a series of Channel-Averaged VSS (CAVSS) blocks and upsampling operations. The architecture of a CAVSS block is shown in \cref{cvass_arch}. It has a VSS block in the beginning that processes the output from the Difference Module. However, while the VSS block can extract global context well, it struggles to learn the inter-channel dependencies \cite{sigma}. Hence, the CAVSS additionally contains average-pool and max-pool operations along the channel dimension, that cater to the channel information, similar to CBAM \cite{cbam}. Thus, the VSS and the Channel-Averaged Pooling layers learn the spatial as well as the channel context. The Mask Decoder follows a UNet \cite{unet} structure, where the output of a Difference Module passes through the CAVSS block, is upsampled and added to the output of the previous Difference Module. Adding these skip connections allows the model to benefit from a wider context and reduce the effect of vanishing gradients. Finally, the output from the last CAVSS block is upsampled to the original image size and fed into a classifier that classifies each pixel as ``Change" or ``No Change" classes.

\section{Experiments and Results}
\noindent\textbf{Datasets: }We compare our method with current state-of-the-art (SOTA) methods for change detection on four widely used public datasets: WHU-CD \cite{whu}, DSIFN-CD \cite{dsifn}, LEVIR-CD \cite{levir}, and CDD \cite{cdd}. For all the datasets, we use the same preprocessing and data splits as \cite{changeformer}.
WHU-CD consists of paired aerial images from 2012 and 2016. It represents an earthquake-affected area, with many changes including rebuilt and new buildings. The training set consists of 5947 image pairs, while the validation set and test set have 743 and 744 pairs respectively.
DSIFN-CD \cite{dsifn} has images from six cities in China from different times. The training, validation and test sets have 14400, 1360, and 192 image pairs respectively.
LEVIR-CD \cite{levir} consists of remote sensing images, collected from Google Earth and majorly covers building appearances and disappearances. It has 7120 image pairs in training set, 1024 in validation set and 2048 in the testing set.
CDD \cite{cdd} consists of Google Earth remote sensing images from varying seasons. It caters to changes in large architectural structures, cars, and seasonal changes in natural vegetation.
All datasets consist of high-resolution images and the data splits were created using $256\times 256$ crops from the original images, and obtained from \cite{changeformer}.\\

\noindent\textbf{Experimental Setup: }We use the AdamW optimizer \cite{adamw} with initial learning rate \(6e^{-5}\) and weight decay \(0.01\). The batch size during training is kept at \(8\) and the model is trained for \(150\) epochs. For initialization of the SIE, we use the ImageNet 1k pretrained model VMamba-Small \cite{vmamba}. The training is divided among four Nvidia RTX A5000 GPUs, each taking ~6GB memory, while just one of these is used during evaluation. Following \cite{ddpmcd}, we use three metrics for evaluation - F1 score, Intersection-Over-Union (IoU) and Overall Accuracy (OA).\\

\noindent\textbf{Results: } The quantitative results comparing M-CD with existing SOTA methods are tabulated in \cref{results_natural}. The first three rows in the table represent methods that employ CNNs for CD. Among these, SNUNet \cite{snunet} and FC-Siam-conc \cite{fcsiam} start training with random weight initialization, while DT-SCN \cite{dtscn} and IFNet \cite{ifnet} use pretrained ImageNet 1k weights for initializing the models. While convolution operations are useful in capturing the spatial context of an image, they lack in comparing the temporal differences between the two images in CD. In contrast, STANet \cite{stanet} uses spatial and temporal attention, while BIT \cite{bit} and ChangeFormer \cite{changeformer} adapt transformers and employ cross-attention operations between the two images, which significantly improves results over convolution-based methods.

Other recent methods like SiamSiam \cite{siamsiam} and SaDL-CD \cite{sadlcd} require a large amount of pertaining and hence, we only present the results which were mentioned in their original papers, namely the F1 score and IoU score on the WHU-CD and LEVIR-CD datasets. We find that our method is able to significantly outperform all the self-supervision-based approaches without requiring a large amount of data. This shows the effectiveness of the Mamaba-based modeling over strong pretrained encoders.
DDPM-CD \cite{ddpmcd} instead learns a diffusion model to learn the structure of remote sensing images and uses its encoder to generate strong predictions, thus being the current SOTA for CD. 

On the other hand, it was shown that Selective State Space Models like Mamba are capable of learning through a larger receptive field as compared to convolutions or transformer-based mechanisms \cite{mamba,vmamba}. This can be seen through the on-par performance of Mamba-based methods like RSMamba \cite{rsmamba}, CDMamba \cite{cdmamba} and ChangeMamba \cite{changemamba}. However, we see that they do not improve the performance significantly. In contrast, by using multi-scale features and a dedicated module to model the temporal difference, we see that our method M-CD significantly improves over existing methods. M-CD is able to outperform CNN-based methods by around 7-10\% in IoU score, and outperform transformer-based methods by around 5\%. In addition, it also performs better than DDPM-CD and Mamba-based methods, with a rise of around 3\% in IoU score. At the same time, our method is also initialized with ImageNet 1k weights and so it does not require a large amount of pretraining with remote sensing image data, unlike DDPM-CD. All results with M-CD have a p-value of at most \(10^{-5}\) compared to other methods, indicating statistical significance.

\begin{figure*}[t!]
  \center
  {\includegraphics[width=.85\linewidth]{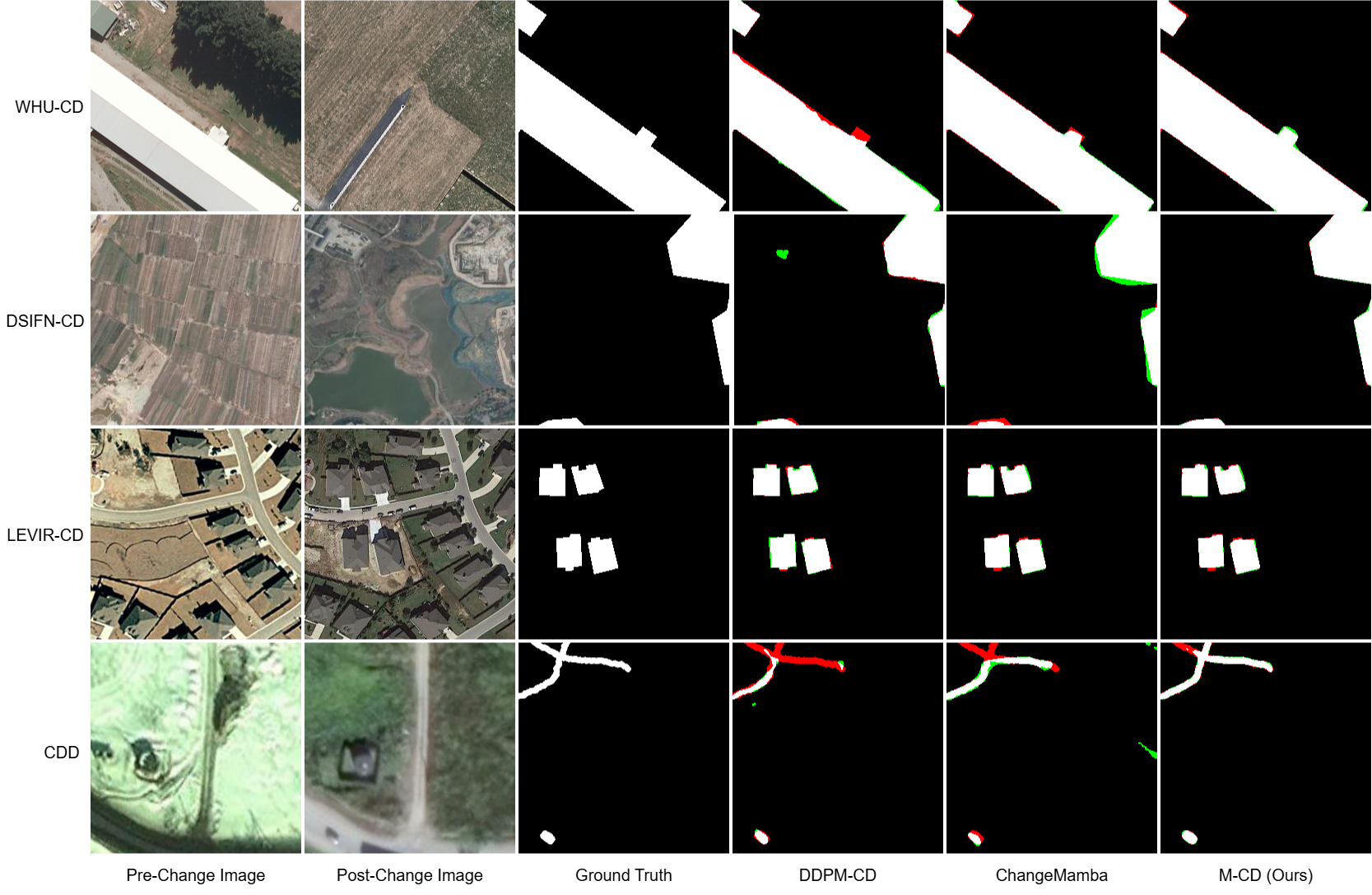}}
\caption{Qualitative results on four public datasets. White represents true positives, black represents true negatives, green represents false positives and red represents false negatives.}  \label{results_fig}
\end{figure*}

We present the qualitative results of our method in \cref{results_fig}. We see that while DDPM-CD generates some artefacts is the output masks, M-CD generates cleaner segmentation predictions, with more precise boundaries.
\begin{table*}
\begin{center}
\resizebox{1.85\columnwidth}{!}{
\begin{tabular}
{@{\extracolsep{4pt}}c c c c c c c c c c c c c c @{}}
\toprule
 & & \multicolumn{3}{c}{WHU-CD \cite{whu}} & \multicolumn{3}{c}{DSIFN-CD}\cite{dsifn} & \multicolumn{3}{c}{LEVIR-CD \cite{levir}} & \multicolumn{3}{c}{CDD \cite{cdd}}\\
\cline{3-5} \cline{6-8} \cline{9-11} \cline{12-14} \\
Method & Extra training data & F1 (\(\uparrow\)) & IoU (\(\uparrow\)) & OA (\(\uparrow\)) & F1 (\(\uparrow\)) & IoU (\(\uparrow\)) & OA (\(\uparrow\)) & F1 (\(\uparrow\)) & IoU (\(\uparrow\)) & OA (\(\uparrow\)) & F1 (\(\uparrow\)) & IoU (\(\uparrow\)) & OA (\(\uparrow\)) \\
\midrule
\textcolor{blue}{CNN-based Methods:} & \multicolumn{13}{l}{}\\
FC-Siam-conc \cite{fcsiam} & None & 0.798 & 0.665 & 98.5 & 0.597 & 0.426 & 87.6 & 0.837 & 0.720 & 98.5 & 0.751 & 0.601 & 94.9\\
SNUNet \cite{snunet}& None & 0.835 & 0.717 & 98.7 & 0.662 & 0.495 & 87.3 & 0.882 & 0.788 & 98.8 & 0.839 & 0.721 & 96.2\\
IFNet \cite{ifnet}& IN1k & 0.834 & 0.715 & 98.8 & 0.601 & 0.430 & 87.8 & 0.881 & 0.788 & 98.9 & 0.840 & 0.719 & 96.03\\
\midrule
\textcolor{blue}{CNN + Attention based Methods:} & \multicolumn{13}{l}{}\\
DT-SCN \cite{dtscn} & IN1k & 0.914 & 0.842 & 99.3 & 0.706 & 0.545 & 82.9 & 0.877 & 0.781 & 98.8 & 0.921 & 0.853 & 98.2\\
STANet \cite{stanet} & IN1k & 0.823 & 0.700 & 98.5 & 0.645 & 0.478 & 88.5 & 0.873 & 0.774 & 98.7 & 0.841 & 0.722 & 96.1\\
\midrule
\textcolor{blue}{Transformer-based Methods:} & \multicolumn{13}{l}{}\\
BIT \cite{bit}& IN1k & 0.905 & 0.834 & 99.3 & 0.876 & 0.780 & 92.3 & 0.893 & 0.807 & 98.92 & 0.889 & 0.800 & 97.5\\
ChangeFormer \cite{changeformer}& None & 0.886 & 0.795 & 99.12 & 0.947 & 0.887 & 93.2 & 0.904 & 0.825 & 99.0 & 0.946 & 0.898 & 98.7\\
\midrule
\textcolor{blue}{Methods with self supervised pretraining} & \multicolumn{13}{c}{}\\
SiamSiam \cite{siamsiam} & IN1k, IBSD, Google Earth & 0.847 & 0.734 & - & - &- & - & 0.880 & 0.786 & - & - & - & -\\
MoCo-v2 \cite{mocov2} & IN1k, IBSD, Google Earth & 0.882 & 0.789 & - & - & - & - & 0.879 & 0.784 & - & - & - & -\\
DenseCL \cite{densecl}& IN1k, IBSD, Google Earth & 0.867 & 0.765 & - & - & - & - & 0.869 & 0.780 & - & - & - & -\\
CMC \cite{cmc} & IN1k, IBSD, Google Earth & 0.886 & 0.795 & - & - & - & - & 0.877 & 0.780 & - & - & - & -\\
SeCo \cite{seco} & IN1k, IBSD, Google Earth & 0.883 & 0.790 & - & - & - & - & 0.881 & 0.787 & - & - & - & -\\
SaDL-CD \cite{sadlcd} & IN1k, IBSD, Google Earth & 0.909 & 0.833 & - & - & - & - & 0.899 & 0.818 & - & - & - & -\\
\midrule
\textcolor{blue}{Mamba-based Methods} & \multicolumn{13}{c}{}\\
RSMamba \cite{rsmamba} & IN1k & 0.927 & 0.865 & 99.4 & 0.965 & 0.913 & 97.0 & 0.897 & 0.814 & 98.9 & 0.943 & 0.902 & 98.8 \\
ChangeMamba \cite{changemamba} & IN1k & 0.925 & 0.861 & 99.4 & 0.875 & 0.778 & 95.8 & 90.16 & 0.821 & 99.0 & 0.944 & \underline{0.920} & 99.0 \\
CDMamba \cite{cdmamba} & IN1k & \underline{0.937} & \underline{0.882} & \underline{99.5} & 0.966 & \underline{0.914} & 97.0 & 0.907 & 0.831 & 99.0 & \underline{0.960} & 0.919 & \underline{99.1} \\
\midrule
\textcolor{blue}{Diffusion-based Methods} & \multicolumn{13}{c}{}\\
DDPM-CD \cite{ddpmcd}& Google Earth & 0.927 & 0.863 & 99.4 & \underline{0.967} & 0.913 & \underline{97.1} & \underline{0.909} & \underline{0.833} & \underline{99.1} & 0.956 & 0.916 & 99.0\\
\midrule
M-CD (Ours) & IN1k & \textbf{0.953} & \textbf{0.911} & \textbf{99.6} & \textbf{0.970} & \textbf{0.935} & \textbf{98.9} & \textbf{0.921} & \textbf{0.850} & \textbf{99.2} & \textbf{0.982} & \textbf{0.963} & \textbf{99.5}\\
\bottomrule
\end{tabular}}
\caption{Comparison of M-CD with respect to SOTA CD methods. F1 denotes the F1 metric, IoU denotes the Intersection-Over-Union metric and OA denotes the overall pixel accuracy. IN1k denotes training data from ImageNet 1k dataset \cite{imagenet1k}. The best result is indicated in \textbf{bold} and the second-best result is \underline{underlined}. Our method outperforms existing methods for all datasets.}
\label{results_natural}
\end{center}
\end{table*}

\section{Ablation Studies}
\noindent\textbf{Varying Mamba Backbones}: In our main experiments, we use the VMamba-Small backbone in the SIE. VMamba \cite{vmamba} provides three different model settings, namely Vmamba Tiny, Small and Base in the increasing order of complexity. We measure the effect of the choice of the backbone for our task by comparing the results on LEVIR-CD in \cref{ablation:backbone}. We find that while there is no significant difference in the performance, VMamba-Small performs slightly better. One reason for this could be that the number of parameters from VMamba-Small to Base increase by 70\%, which can be more prone to overfitting. Hence, there is no significant improvement, when going from Small to Tiny. A similar argument can also hold while going from Tiny to Small.

\begin{table}
\begin{center}
\resizebox{0.7\columnwidth}{!}{
\begin{tabular}
{@{\extracolsep{4pt}}c c c @{}}
\toprule
Backbone & Number of Parameters & LEVIR-CD IoU \\
\midrule
Tiny & 36.5 M & 0.848 \\
Small & 58.1 M & 0.850 \\
Base & 100.7 M & 0.845 \\
\bottomrule
\end{tabular}}
\caption{Ablation Experiments with varying encoder backbones. While the number of parameters increase significantly, there is no significant change in model performance on LEVIR-CD.}
\label{ablation:backbone}
\end{center}
\end{table}

\noindent\textbf{Varying Difference Module: }We measure the effect of the design choice of the Difference Module (DM) on model performance for LEVIR-CD and DSIFN-CD datasets in \cref{ablation:dm}. We evaluate three different designs for the Difference Module. In the first row, ``Difference" signifies that the DM simply subtracts the post-change image from the pre-change image and sends the difference to the decoder. While this approach seems naive, this allows the decoder to learn from the residuals between the image features and performs well on the Change Detection task. In the second row, we use the strategy of combining of Cross Mamba and Concat Mamba blocks used by Sigma \cite{sigma} to generate segmentation masks using multiple modalities. This technique performs well when multiple modalities guide the model towards a single segmentation results. However, for the case of Change Detection, we find that interchanging the parameters for the two images during training, as proposed by Sigma leads to a lower model performance. This could be because an interchange encourages the model to treat the post-change image as being similar to the pre-change image semantically. This proves to be beneficial when fusing two modalities like in Sigma, but is harmful for the task of Change Detection. Hence, we see a lower performance. Finally, the last row represents our method, which uses the concatenation of the feature vectors from the two images. This allows the DM to combine the two features with more freedom than simple difference, while not enforcing similarity. Hence, it achieves a higher performance than the other two strategies.

\begin{table}
\begin{center}
\resizebox{0.75\columnwidth}{!}{
\begin{tabular}
{@{\extracolsep{4pt}}c c c@{}}
\toprule
Difference Module Design Strategy & LEVIR-CD IoU & DSIFN-CD IoU\\
\midrule
Difference & 0.840 & 0.924 \\
Cross+Concatenation \cite{sigma} & 0.821 & 0.909 \\
Concatenation (Ours) & \textbf{0.850} & \textbf{0.935} \\
\bottomrule
\end{tabular}}
\caption{Ablation Experiments with varying DM designs. Our method (concatenation) shows higher performance on the LEVIR-CD and DSIFN-CD datasets over other strategies like difference of features, or combining parameter crossing and concatenation, as proposed in \cite{sigma}.}
\label{ablation:dm}
\end{center}
\end{table}

\noindent\textbf{Computational Complexity Analysis: }We calculate the compute required for running our method using three metrics, namely the number of trainable parameters the model has, amount of floating point operations required for a forward pass, in gigaflops (GFLOPS) and the average inference time required for one pair of pre and post-change image inputs. The image size for these experiments is 256X256 and one Nvidia RTX A5000 GPU is used for the calculation of these metrics. The results are tabulated in \cref{ablation:cc}. We see that our method has a greater number of trainable parameters as well as has a higher inference time as compared to other SOTA methods. However, GFLOPS in Mamba-based architectures have a linear scaling relation with the number of parameters \cite{mamba}. Hence, the number of floating point operations are comparable to other methods and significantly lower than DDPM-CD. For rest of the methods, while the computational complexity of our method is greater, the gain in performance, especially in the absence of large amounts of pertaining, outweighs this limitation in our opinion. 

\begin{table}
\begin{center}
\resizebox{\columnwidth}{!}{
\begin{tabular}
{@{\extracolsep{4pt}}c c c c@{}}
\toprule
Method & Trainable Parameters (million) & GFLOPS & Average Inference Time Per Image Pair (ms)\\
\midrule
SiamSiam & 12.49 & 4.76 & 1.04\\
MoCo-v2 & 11.24 & 4.76 & 1.92\\
DenseCL & 11.69 & 4.76 & 2.66\\
CMC & 22.48 & 4.66 & 1.55\\
SeCo & 12.16 & 9.52 & 3.62\\
DDPM-CD \cite{ddpmcd} & 46.41 & 2175.46 & 88.10\\
M-CD (Ours) & 69.80 & 29.58 & 160\\
\bottomrule
\end{tabular}}
\caption{Computational Complexity Analysis of M-CD. Our method has greater trainable parameters and inference time. The number of GFLOPs are comparable. However, while this is a limitation of M-CD, we believe the improvement in performance without additional training data outweighs the shortcomings.}
\label{ablation:cc}
\end{center}
\end{table}

\begin{figure}[htp!]
  \center
{\includegraphics[width=0.7\linewidth]{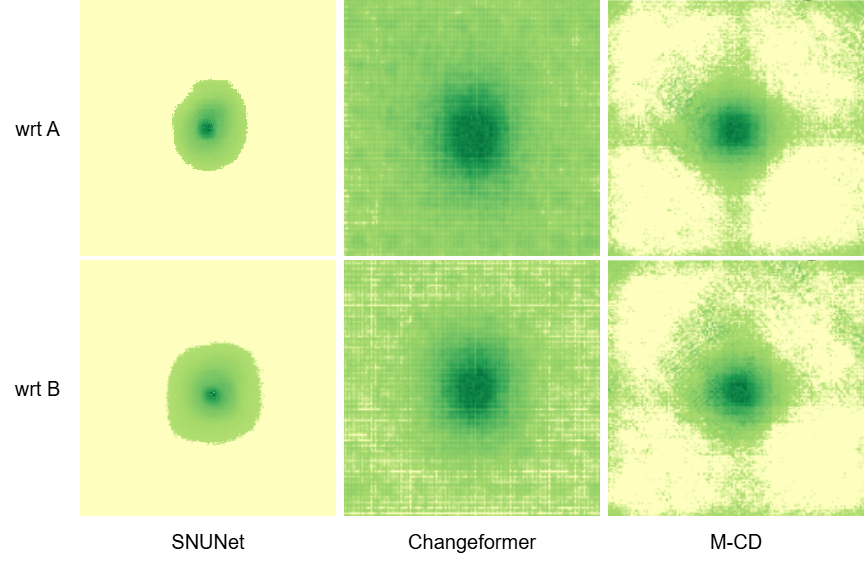}}
\caption{Effective Receptive Field (ERF) of SNUNet v.s. ChangeFormer v.s. M-CD for WHU dataset. The top row indicates ERF with respect to the pre-change image (A) and the bottom row indicates ERF with respect to the post-change image (B). Darker shade of green indicates higher dependence on the input pixel. M-CD and Transformer have a more global ERF than SNUNet. M-CD has a more structured ERF than the uniform ERF of ChangeFormer}  \label{erf}
\end{figure}
\noindent\textbf{Effective Receptive Field (ERF): }We utilize the code provided by VMamba \cite{vmamba} to generate the ERF for our method (Mamba-based) in comparison to ChangeFormer \cite{changeformer} (transformer-based) and SNUNet \cite{snunet} (CNN-based). This represents the importance of every input image pixel for generating the center of the predicted mask. As shown in \cref{erf}, using Mamba architecture includes information from both the input images at a more global level compared to CNNs, as seen from the darker shades of green at non-centric parts of the ERF. On the other hand, the transformer has a uniformly distributed attention map unlike Mamba, which is more structured. This shows that Mamba can achieve a wider receptive field across both the pre and post-change images, and can find and prioritize certain areas of the inputs which are more relevant for CD. These results are consistent with similar studies conducted in \cite{vmamba}.

\section{Conclusion}
In this work, we introduce M-CD, a model for the task of Change Detection, that is based upon the Selective State Space (a.k.a. Mamba) architecture. M-CD comprises of a Siamese Image Encoder to featurize the input images, a Difference Module that performs multi-scale analysis of the change of interest, and finally, a Mask Decoder to generate the segmentation masks. Extensive experiments on four widely used datasets clearly show that M-CD beats the existing SOTA methods. At the same time, it requires much lesser pretaining than the recently proposed self-supervised or diffusion methods, while outperforming them significantly, thus making M-CD the new SOTA in this field.

{\small
\bibliographystyle{ieee_fullname}
\bibliography{egbib}
}

\end{document}